\documentclass[11pt,a4paper]{article}
\usepackage[hyperref]{emnlp2018}
\usepackage{times}
\usepackage{latexsym}

\usepackage{url}
\usepackage{graphicx}
\usepackage{caption} 
\aclfinalcopy 

\title{The Effect of Context on Metaphor Paraphrase Aptness Judgments}

\author{Yuri Bizzoni \\
  University of Gothenburg\\
  {\tt yuri.bizzoni@gu.se} \\\And
  Shalom Lappin\\
  University of Gothenburg\\
  {\tt shalom.lappin@gu.se}}

\date{}

\begin{document}
\maketitle
\begin{abstract}
  We conduct two experiments to study the effect of context on metaphor paraphrase aptness judgments. The first is an AMT crowd source task in which speakers rank metaphor-paraphrase candidate sentence pairs in short document contexts for paraphrase aptness. In the second we train a composite DNN to predict these human judgments, first in binary classifier mode, and then as gradient ratings. We found that for both mean human judgments and our DNN's
predictions, adding document context compresses the aptness scores towards the center of the scale, raising low out-of-context ratings and decreasing high out-of-context scores. We offer a provisional explanation for this compression effect.
\end{abstract}

\section{Introduction}

A metaphor is a way of forcing the normal boundaries of a word's meaning in order to better express an experience, a concept or an idea.  To a native speaker's ear some metaphors sound more conventional (like the usage of the words \textit{ear} and \textit{sound} in this sentence), others more original. This is not the only dimension along which to judge a metaphor. One of the most important qualities of a metaphor is its appropriateness, its \textit{aptness}: how good is a metaphor for conveying a given experience or concept. While a metaphor's degree of conventionality can be measured through probabilistic methods, like language models, it is harder to represent its aptness.
\citet{chiappe2003aptness} define \textit{aptness} as ``the extent to which a comparison captures important features of the topic".

It is possible to express an opinion about some metaphors' and similes' aptness (at least to a degree) without previously knowing what they are trying to convey, or the context in which they appear\footnote{While it can be argued that metaphors and similes at 
  some level work differently and cannot always be  
  considered as variations of the same phenomenon 
  \citep{sam2006relation,glucksberg2008metaphors}, for this study we treat 
  them as belonging to the same category of figurative
  language.}. 
For example, we don't need a particular context or frame of reference to construe the simile \textit{She was screaming like a turtle} as strange, and less apt for expressing the quality of a scream than \textit{She was screaming like a banshee}.
In this case, the reason why the simile in the second sentence works best is intuitive. A salient characteristic of a banshee is a powerful scream. Turtles are not known for screaming, and so it is harder to define the quality of a scream through such a comparison, except as a form of irony.\footnote{It is important not to confuse
  aptness with transparency. The latter measures how easy it is to 
  understand a comparison. \citet{chiappe2003aptness} claim, for 
  example, that many literary or poetic metaphors score high on
  aptness and low on transparency, in that they capture the nature of 
  the topic very well, but it is not always clear why they work.}
Other cases are more complicated to decide upon. The simile \textit{crying like a fire in the sun} (\emph{It's All Over Now, Baby Blue}, Bob Dylan) is powerfully apt for many readers, but simply odd for others. Fire and sun are not known to cry in any way. But at the same time the simile can capture the association we draw between something strong and intense in other senses - vision, touch, etc. - and a loud cry. 

Nonetheless, most metaphors and similes need some kind of context, or external reference point to be interpreted. The sentence \textit{The old lady had a heart of stone} is apt if the old lady is cruel or indifferent, but it is inappropriate as a description of a situation  in which the old lady is kind and caring. We assume that, to an average reader's sensibility, the sentence models the situation in a satisfactory way only in the first case.

This is the approach to metaphor aptness that we assume in this paper.
Following 
\citet{bizzoni2018}, 
we treat a metaphor as apt in relation to a literal expression that it paraphrases.\footnote{
\citet{bizzoni2018} 
apply 
 \citet{Bizzoni&Lappin2017}'s modeling work on general
 paraphrase to metaphor.} 
If the metaphor is judged to be a good paraphrase, then it closely expresses the core information of the literal sentence through its metaphorical shift. We refer to the prediction of readers' judgments on the aptness candidates for the literal paraphrase of a metaphor as the \emph{metaphor paraphrase aptness task} (MPAT).
{
\citet{bizzoni2018} 
address the MPAT by using Amazon Mechanical Turk (AMT) to obtain crowd sourced annotations of metaphor-paraphrase candidate pairs. They train a composite Deep Neural Network (DNN) on a portion of their annotated corpus, and test it on the remaining part. Testing involves using the DNN as a binary classifier on paraphrase candidates. They derive predictions of gradient paraphrase aptness for their test set, and assess them by Pearson coefficient correlation to the mean judgments of their crowd sourced annotation of this set. 
Both training and testing are done independently of any document context for the metaphorical sentence and its literal paraphrase candidates.

In this paper we study the role of context on readers' judgments concerning the aptness of metaphor paraphrase candidates. We look at 
the accuracy of {
\citet{bizzoni2018}'s 
DNN when trained and tested on contextually embedded metaphor-paraphrase pairs for the MPAT.
In Section \ref{annotation} we describe an AMT experiment in which annotators judge metaphors and paraphrases embodied in small document contexts, and in Section \ref{annotation_results} we discuss the results of this experiment.
In Section \ref{modelling} we describe our MPAT modeling experiment,
and in Section \ref{modelling_results} we discuss the results of this experiment.
Section \ref{related_work} briefly surveys some related work. In Section \ref{conclusions} we draw conclusions from our study, and we indicate directions for future work in this area.

\section{Annotating Metaphor-Paraphrase Pairs in Contexts}
\label{annotation}

\citet{bizzoni2018} 
have recently produced a dataset of paraphrases containing metaphors designed to allow both supervised binary classification and gradient ranking. This dataset contains several pairs of sentences, where in each pair the first sentence contains a metaphor, and the second is a literal paraphrase candidate.

This corpus was constructed with a view to representing a large variety of syntactic structures and semantic phenomena in metaphorical sentences. Many of these structures and phenomena do not occur as metaphorical expressions, with any frequency, in natural text and were therefore introduced through hand crafted examples. 

Each pair of sentences in the corpus has been rated by AMT annotators for paraphrase aptness on a scale of 1-4, with 4 being the highest degree of aptness. 
In 
{
\citet{bizzoni2018}'s 
dataset, sentences come in groups of five, where the first element is the ``reference element" with a metaphorical expression, and the remaining four sentences are ``candidates" that stand in a degree of paraphrasehood to the reference. 

Here is an example of a metaphor-paraphrase candidate pair.

\begin{enumerate}
\item[1a.] The crowd was a roaring river.
\item[b.] The crowd was huge and noisy. 
\end{enumerate}
The average AMT paraphrase score for this pair is 4.0, indicating a high degree of aptness. 

We extracted 200 sentence pairs from 
{
\citet{bizzoni2018}'s 
dataset and provided each pair with a document context consisting of a preceding and a following sentence\footnote{Our annotated data set and the 
  code for our model is available at \url{https://github.com/yuri-bizzoni/Metaphor-Paraphrase} .}, as in the following example.

\begin{enumerate}
\item[2a.] They had arrived in the capital city. \textbf{The crowd was a roaring river}. It was glorious.
\item[b.] They had arrived in the capital city. \textbf{The crowd was huge and noisy}. It was glorious. 
\end{enumerate}

One of the authors constructed most of these contexts by hand. In some cases, it was possible to locate the original metaphor in an existing document. This was the case for 
\begin{enumerate}
\item[(i)] Literary metaphors extracted from poetry or novels, and
\item[(ii)] Short conventional metaphors (\textit{The President brushed aside the accusations}, \textit{Time flies}) that can be found, with small variations, in a number of texts.
\end{enumerate}

For these cases, a variant of the existing context was added to both the metaphorical and the literal sentences. We introduced small modifications to keep the context short and clear, and to avoid copyright issues. We lightly modified the contexts of metaphors extracted from corpora when the original context was too long, ie. when the contextual sentences of the selected metaphor were longer than the maximum length we specified for our corpus. In such cases we reduced the length of the sentence, while sustaining its meaning.

The context was designed to sound as natural as possible. Since the same context is used for metaphors and their literal candidate paraphrases, we tried to design short contexts that make sense for both the figurative and the literal sentences, even when the pair had been judged as non-paraphrases. We kept the context as neutral as possible in order to avoid a distortion in crowd source ratings.

For example, in the following pair of sentences, the literal sentence is \textit{not} a good paraphrase of the figurative one (a simile).

\begin{enumerate}
\item[3a.] He is grinning like an ape.
\item[b.] He is smiling in a charming way. (\textit{average score: } 1.9)
\end{enumerate}

We opted for a context that is natural for both sentences.

\begin{enumerate}
\item[4a.] Look at him. \textbf{He is grinning like an ape.} He feels so confident and self-assured.
\item[b.] Look at him. \textbf{He is smiling in a charming way.} He feels so confident and self-assured. 
\end{enumerate}

We sought to avoid, whenever possible, an incongruous context for one of the sentences that could influence our annotators' ratings. 

We collected a sub-corpus of 200 contextually embedded pairs of sentences. We tried to keep our data as balanced as possible, drawing from all four rating classes of paraphrase aptness ratings (between 1 to 4) that 
{
\citet{bizzoni2018} 
obtained. We selected 44 pairs of \textit{1} ratings, 51 pairs of \textit{2}, 43 pairs of \textit{3} and 62 pairs of \textit{4}. 

We then used AMT crowd sourcing to rate the contextualized  paraphrase pairs, so that we could observe the effect of document context on assessments of metaphor paraphrase aptness. 

To test the reproducibility of 
{
\citet{bizzoni2018}'s 
ratings, we launched a pilot study for 10 original non-contextually embedded pairs, selected from all four classes of aptness. We observed that the annotators provided mean ratings very similar to those reported in 
{
\citet{bizzoni2018}. 
The Pearson coefficent correlation between the mean judgments of our out-of-context pilot annotations and 
{
\citet{bizzoni2018}'s 
annotations for the same pair was over 0.9.
We then conducted an AMT annotation task for the 200 contextualised pairs. On average, 20 different annotators rated each pair. We considered as ``rogue" those annotators who rated the large majority of pairs with very high or very low scores, and those who responded inconsistently to two ``trap" pairs. After filtering out the rogues, we had an average of 14 annotators per pair.

\section{Annotation Results}
\label{annotation_results}

We found a Pearson correlation of 0.81 between the in-context and out-of-context mean human paraphrase ratings for our two corpora. This correlation is virtually identical to the one that \citet{lappin2018} report for mean acceptability ratings of out-of-context to in-context sentences in their crowd source experiment. It is interesting that a relatively high level of ranking correspondence should occur in mean judgments for sentences presented out of and within document contexts, for two entirely distinct tasks.  

Our main result concerns the effect of context on mean paraphrase judgment. We observed that it tends to flatten aptness ratings towards the center of the rating scale. 71.1\% of the metaphors that had been considered highly apt (average rounded score of \textit{4}) in the context-less pairs received a more moderate judgment (average rounded score of \textit{3}), but the reverse movement was rare. Only 5\% of pairs rated \textit{3} out of context (2 pairs) were boosted to a mean rating of \textit{4} in context. At the other end of the scale, 68.2\% of the metaphors judged at \textit{1} category of aptness out of context were raised to a mean of \textit{2} in context, while only the 3.9\%  of pairs rated \textit{2} out of context were lowered to \textit{1} in context. 

Ratings at the middle of the scale - \textit{2} (defined as semantically related non-paraphrases) and \textit{3} (imperfect or loose paraphrases) - remained largely stable, with little movement in either direction. 9.8\% of pairs rated \textit{2} were re-ranked as \textit{3} when presented in context, and 10\% of pairs ranked at \textit{3} changed to \textit{2}. 
The division between \textit{2} and \textit{3}  separates paraphrases from non-paraphrases. Our results suggest that this binary rating of paraphrase aptness was not strongly affected by context. Context  operates at the extremes of our scale, raising low aptness ratings and lowering high aptness ratings. This effect is clearly indicated in the regression chart in Fig \ref{regression_chart}.

\begin{figure*}[h]
\begin{center}
\includegraphics[scale =.7]{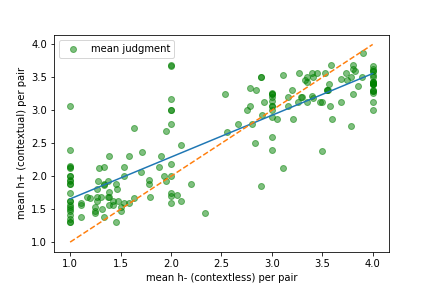}
\caption{In-context and out-of-context mean
ratings. Points above the broken diagonal line represent sentence pairs which received a higher rating when presented in context. The total least-square linear regression is shown as the second line.}
\label{regression_chart}
\end{center}
\end{figure*}


This effect of context on human ratings is very similar to the one reported in \citet{lappin2018}. They find that sentences rated as ill formed  out of context are improved when they are presented in their document contexts. However the mean ratings for sentences judged to be highly acceptable out of context declined when assessed in context. \citet{lappin2018}'s linear regression chart for the correlation between out-of-context and in-context acceptability judgments looks remarkably like our Fig \ref{regression_chart}. There is, then, a striking parallel in the compression pattern that context appears to exert on human judgments for two entirely different linguistic properties. 

This pattern requires an explanation. \citet{lappin2018} suggest that adding context causes speakers to focus on broader semantic and pragmatic issues of discourse coherence, rather than simply judging syntactic well formedness (measured as naturalness) when a sentence is considered in isolation. On this view, compression of rating results from a pressure to construct a plausible interpretation for any sentence within its context. 

If this is the case, an analogous process may generate the same compression effect for metaphor aptness assessment of sentence pairs in context. Speakers may attempt to achieve broader discourse coherence when assessing the metaphor-paraphrase aptness relation in a document context. Out of context they focus more narrowly on the semantic relations between a metaphorical sentence and its paraphrase candidate. Therefore, this relation is at the centre of a speaker's concern, and it receives more fine-grained assessment when considered out of context than in context. This issue clearly requires further research.

\section{Modelling Paraphrase Judgments in Context}
\label{modelling}

We use the DNN model described in 
{
\citet{bizzoni2018} 
to predict aptness judgments for in-context paraphrase pairs. It has three main components: 

\begin{enumerate}
\item Two encoders that learn the representations of two sentences separately
\item  A unified layer that merges the output of the encoders
\item A final set of fully connected layers that operate on the merged representation of the two sentences to generate a judgment.
\end{enumerate}   

The encoder for each pair of sentences taken as input is composed of two parallel "Atrous" Convolutional Neural Networks (CNNs) and LSTM RNNs, feeding two sequenced fully connected layers. 

The encoder is preloaded with the lexical embeddings from Word2vec \cite{Mikolov&etal2013}. The sequences of word embeddings that we use as input provides the model with dense word-level information, while the model tries to generalize over these embedding patterns. 

The combination of a CNN and an LSTM allows us to capture both long-distance syntactic and semantic relations, best identified by a CNN, and the sequential nature of the input, most efficiently identified by an LSTM. Several existing studies, cited in \citet{Bizzoni&Lappin2017}, demonstrate the advantages of combining CNNs and LSTMs to process texts. 

The model produces a single classifier value between 0 and 1. We transform this score into a binary output of 0 or 1 by applying a threshold of 0.5 for assigning 1.

The architecture of the model is given in Fig \ref{model}.

\begin{figure*}
\begin{center}
\includegraphics[width=70mm,scale=.9]{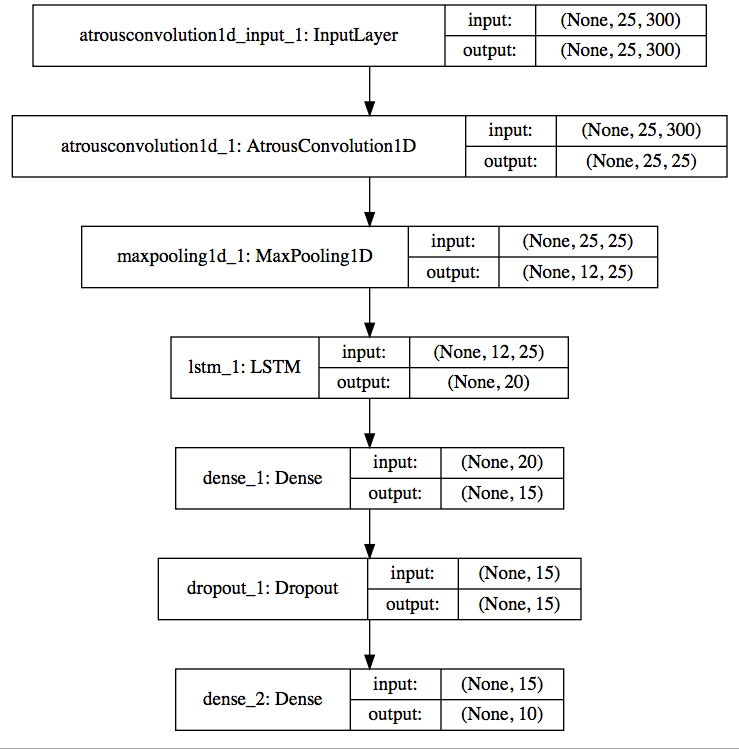}
\caption{DNN encoder for predicting metaphorical paraphrase aptness from \citet{bizzoni2018}. Each encoder represents a sentence as a 10-dimensional vector. These vectors are concatenated to compute a single score for the pair of input sentences. 
}
\label{model}
\end{center}
\end{figure*}

We use the same general protocol as 
\citet{bizzoni2018} 
for training with supervised learning, and testing the model. 

Using 
\citet{bizzoni2018}'s 
out-of- context metaphor dataset and our contextualized extension of this set, we apply four variants of the training and testing protocol. 

\begin{enumerate}
\item Training and testing on the in-context dataset.
\item Training on the out-of-context dataset, and testing on the in-context dataset.
\item Training on the in-context dataset, and testing on the out-of-context dataset.
\item Training and testing on the out-of-context dataset
(
\citet{bizzoni2018}'s 
original experiment provides the results for out-of-context training and testing).
\end{enumerate}

When we train or test the model on the out-of-context dataset, we use 
{
\citet{bizzoni2018}'s 
original annotated corpus of 800 metaphor-paraphrase pairs. The in-context dataset contains 200 annotated pairs. 

\section{MPAT Modelling Results}
\label{modelling_results}

We use the model both to predict binary classification of a metaphor paraphrase candidate, and to generate gradient aptness ratings on the 4 category scale (see 
{
\citet{bizzoni2018} 
for details). A positive binary classification is accurate if it is $\geq$ a 2.5 mean human rating. The gradient predictions are derived from the softmax distribution of the output layer of the model.
The results of our modelling experiments are given in Table \ref{model_table}.

\begin{table*}[t!]
\small
\centering
\begin{tabular}{|p{3.0cm}|p{3.0cm}|p{1.5cm}|p{2.0cm}|}
\hline \bf Training set & \bf Test set & \bf F-score & \bf Correlation \\ \hline
With-context* &  With-context* & 0.68  & -0.01\\
Without-context & With-context  &  \textbf{0.72} & \textbf{0.3} \\
With-context & Without-context & 0.6 & 0.02 \\ \hline
Without-context & Without-context & 0.74 & 0.75\\
\hline
\end{tabular}
\vspace{5pt}
\caption{F-score binary classification accuracy and Pearson correlation for three different regimens of supervised learning. The * indicates results for a set of 10-fold cross-validation runs. This was necessary in the first case, when training and testing are both on our small corpus of in-context pairs. In the second and third rows, since we are using the full out-of-context and in-context dataset, we report single-run results. The fourth row is 
\citet{bizzoni2018}'s 
best run result. (Our single-run best result for the first row is an F-score of 0.8 and a Pearson correlation 0.16).}
\label{model_table}
\end{table*}

The main result that we obtain from these experiments is that the model learns binary classification to a reasonable extent on the \textit{in-context} dataset, both when trained on the same kind of data (in-context pairs), and when trained on 
\citet{bizzoni2018}'s 
original dataset (out-of-context pairs). However, the model does not perform well in predicting gradient in-context judgments when trained on in-context pairs. It improves slightly for this task when trained on out-of-context pairs. 

By contrast, it does well in predicting both binary and gradient ratings when trained and tested on out-of-context data sets. 

\citet{lappin2018} also note a decline in Pearson correlation for their DNN models on the task of predicting human in-context acceptability judgments, but it is less drastic. They attribute this decline to the fact that the compression effect renders the gradient judgments less separable, and so harder to predict. 
A similar, but more pronounced version of this effect may account for the difficulty that our model encounters in predicting gradient in-context ratings. The binary classifier achieves greater success for these cases because its training tends to polarise the data in one direction or the other. 

We also observe that the best combination seems to consist in training our model on the original out-of-context dataset and testing it on the in-context pairs. In this configuration we reach an F-score (0.72) only slightly lower than the one reported in 
\citet{bizzoni2018} 
(0.74), and we record the highest Pearson correlation, 0.3 (which is still not strong, compared to 
\citet{bizzoni2018}'s best run,
0.75\footnote{It is also important to consider that their ranking scheme is different from ours: the Pearson correlation reported there is the average of the correlations over all groups of 5 sentences present in the dataset.}). This result may partly be an artifact of the the larger amount of training data provided by the out-of-context pairs.

We can use this variant (out-of-context training and in-context testing) to perform a fine-grained comparison of the model's predicted ratings for the same sentences in and out of context.
When we do this, we observe that out of 200 sentence pairs, our model scores the majority (130 pairs) higher when processed in context than out of context.  A smaller but significant group (70 pairs) receives a lower score when processed in context. The first group's average score \textit{before adding context} (0.48) is consistently lower than that of the second group (0.68). 
Also, as Table \ref{model_table2} indicates, the pairs that our model rated, \textit{out of context}, with a score lower than 0.5 (on the model's softmax distribution), received on average a higher rating \textit{in context}, while the opposite is true for the pairs rated with a score higher than 0.5. 
In general, sentence pairs that were rated highly out of context receive a lower score in context, and vice versa. 
When we did linear regression on the DNNs in and out of context predicted scores, we observed substantially the same compression pattern exhibited by our AMT mean human judgments.  Figure \ref{regression_chart2} plots this regression graph.


\begin{table*}[t!]
\small
\centering
\begin{tabular}{|p{1.5cm}|p{2cm}|p{2cm}|p{1.5cm}|p{1.5cm}|p{1.5cm}|}
\hline \bf OOC score & \bf Number of elements &\bf OOC Mean & \bf OOC Std & \bf IC Mean &\bf IC Std \\ \hline
\bf 0.0-0.5  & 112 &  0.42 & 0.09& 0.54 & 0.1 \\
\bf 0.5-1.0  & 88 & 0.67 &  0.07 & 0.64 & 0.07\\
\hline
\end{tabular}
\vspace{5pt}
\caption{We show the number of pairs that received a low score out of context (first row) and the number of pairs that received a high score out of context (second row). We report the mean score and standard deviation (Std) of the two groups when judged out of context (OOC) and when judged in context (IC) by our model. The model's scores range between 0 and 1. As can be seen, the mean of the low-scoring group rises in context, and the mean of the high-scoring group decreases in context.}
\label{model_table2}
\end{table*}

\begin{figure*}[h]
\begin{center}
\includegraphics[scale=.7]{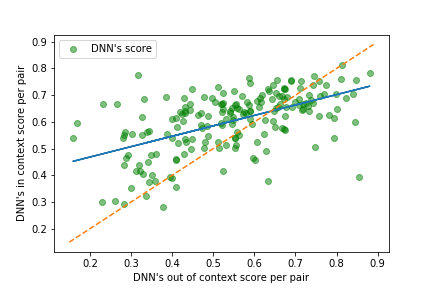}
\caption{In-context and out-of-context ratings assigned by our trained model. Points above the broken diagonal line represent sentence pairs which received a higher rating when presented in context. The total least-square linear regression is shown as the second line.}
\label{regression_chart2}
\end{center}
\end{figure*}




\section{Related Cognitive Work on Metaphor Aptness}
\label{related_work}

\citet{tourangeau1981aptness} present ratings of aptness and comprehensibility for 64 metaphors from two groups of subjects. 
They note that metaphors were perceived as more apt and more comprehensible to the extent that their terms occupied similar positions within dissimilar domains. 
Interestingly, \citet{fainsilber1984does} also present experimental results to claim that imagery does not clearly correlate with metaphor aptness.
Aptness judgments are also subjected to individual differences. 

\citet{blasko1999only} points to such individual differences in metaphor processing. She asked 27 participants to rate 37 metaphors for difficulty, aptness and familiarity, and to write one or more interpretations of the metaphor. Subjects with higher working memory span were able to give more detailed and elaborate interpretations of metaphors. Familiarity and aptness correlated with both high and low span subjects.  For high span subjects aptness of metaphor positively correlated with number of interpretations, while for low span subjects the opposite was true. 

\citet{mccabe1983conceptual} analyses the aptness of metaphors with and without extended context. She finds that domain similarity correlates with aptness judgments in isolated metaphors, but not in \textit{contextualized} metaphors. She also reports that there is no clear correlation between metaphor aptness ratings in isolated and in contextualized examples.
\citet{chiappe2003aptness} study the relation between aptness and comprehensibility in metaphors and similes. They provide experimental results indicating that aptness is a better predictor than comprehensibility for the ``transformation" of a simile into a metaphor. Subjects tended to remember similes as metaphors (i.e. remember \textit{the dancer's arms moved like startled rattlesnakes} as \textit{the dancer's arms \textbf{were} startled rattlesnakes}) if they were judged to be particularly apt, rather than particularly comprehensible. They claim that context might play an important role in this process. They suggest that context should ease the transparency and increase the  aptness of both metaphors and similes. 

\citet{tourangeau1991interpreting} present a series of experiments indicating that metaphors tend to be interpreted through emergent features that were not rated as particularly relevant, either for the tenor or for the vehicle of the metaphor. The number of emergent features that subjects were able to draw from a metaphor seems to correlate with their aptness judgments. 

\citet{bambini2018time} use Event-Related Brain Potentials (ERPs) to study the temporal dynamics of metaphor processing in reading literary texts. They emphasize the influence of context on the ability of a reader to smoothly interpret an unusual metaphor. 

\citet{bambini2016disentangling} use electrophysiological experiments to try to disentangle the effect of a metaphor from that of  its context. They find that de-contextualized metaphors elicited two different brain responses, $N400$ and $P600$, while contextualized metaphors only produced the $P600$ effect. They attribute the $N400$ effect, often observed in neurological studies of metaphors, to expectations about upcoming words in the absence of a predictive context that ``prepares" the reader for the metaphor. They suggest that the $P600$ effect reflects the actual interpretative processing of the metaphor.  

This view is supported by several neurological studies showing that the $N400$ effect arises with unexpected elements, like new presuppositions introduced into a text in a way not implied by the context \cite{masia2017presupposition}, or unexpected associations with a noun-verb combination, not indicated by previous context (for example preceded by neutral context, as in \citet{cosentino2017time}).


\section{Conclusions and Future Work}
\label{conclusions}

We have observed that embedding metaphorical sentences and their paraphrase candidates in a document context generates a compression effect in human metaphor aptness ratings. Context seems to mitigate the perceived aptness of metaphors in two ways. Those metaphor-paraphrase pairs given very low scores out of context receive increased scores in context, while those with very high scores out of context decline in rating when presented in context. At the same time, the demarcation line between paraphrase and non-paraphrase is not particularly affected by the introduction of extended context.

As previously observed by \citet{mccabe1983conceptual}, we found that context has an influence on human aptness ratings for metaphors, although, unlike her results, we did find a correlation between the two sets of ratings. 
\citet{chiappe2003aptness}'s expectation that context should facilitate a metaphor's aptness was supported only in one sense. Aptness increases for low-rated pairs. But it decreases for high-rated pairs. 

We applied 
\citet{bizzoni2018}'s
DNN for the MAPT to an in-context test set, experimenting with both out-of-context and in-context training corpora. We obtained reasonable results for binary classification of paraphrase candidates for aptness, but the performance of the model declined sharply for the prediction of human gradient aptness judgments, relative to its performance on a corresponding out-of-context test set. This appears to be the result of the increased difficulty in separating rating categories introduced by the compression effect.

Strikingly, the linear regression analyses of human aptness judgments for in- and out-of-context paraphrase pairs, and
of our DNN's predictions for these pairs reveal similar compression patterns. These patterns produce ratings that cannot be clearly separated along a linear ranking scale. 

To the best of our knowledge ours is the first study of the effect of context on metaphor aptness on a corpus of this dimension, using crowd sourced human judgments as the gold standard for assessing the predictions of a computational model of paraphrase. We also present the first comparative study of both human and model judgments of metaphor paraphrase for in-context and out-of-context variants of metaphorical sentences.

Finally, the compression effect that context induces on paraphrase judgments corresponds closely to the one observed independently in another task, which is reported in \citet{lappin2018}. We regard this effect as a significant discovery that increases the plausibility and the interest of our results. The fact that it appears clearly with two tasks involving different sorts of DNNs and distinct learning regimes (unsupervised learning with neural network language models for the acceptability prediction task discussed, as opposed to supervised learning with our composite DNN for paraphrase prediction) reduces the likelihood that this effect is an artefact  of our experimental design. 

While our dataset is still small, we are presenting an initial investigation of a phenomenon which is, to date, little studied. We are working to enlarge our dataset and in future work we will expand both our in- and out-of-context annotated metaphor-paraphrase corpora. 

While the corpus we used contains a number of hand crafted examples, it would be preferable to find these example types in natural corpora, and we are currently working on this. We will be extracting a dataset of completely natural (corpus-driven) examples. We are seeking to expand the size of the data set to improve the reliability of our modelling experiments.

We will also experiment with alternative DNN architectures for the MAPT. We will conduct qualitative analyses on the kinds of metaphors and similes that are more prone 
to a context-induced rating switch.

One of our main concerns in future research will be to achieve a better understanding of the compression effect of context on human judgments and DNN models.

 \bibliography{emnlp2018}
 \bibliographystyle{acl_natbib_nourl}

\end{document}